# Mirroring the Parking Target: An Optimal-Control-Based Parking Motion Planner with Strengthened Parking Reliability and Faster Parking Completion*

Jia Hu, *Senior Member, IEEE,* Yongwei Feng, Shuoyuan Li, Haoran Wang, *Member, IEEE,* Jaehyun So, and Junnian Zheng

*Abstract*—Automated Parking Assist (APA) systems are now facing great challenges with low adoption in applications, due to users' concerns about parking capability, reliability, and completion efficiency. To upgrade the conventional APA planners and enhance user's acceptance, this research proposes an optimal-control-based parking motion planner. Its highlight lies in its control logic: planning trajectories by mirroring the parking target. This method enables: i) parking capability in narrow spaces; ii) better parking reliability by expanding Operation Design Domain (ODD); iii) faster completion of parking process; iv) enhanced computational efficiency; v) universal to all types of parking. A comprehensive evaluation is conducted. Results demonstrate the proposed planner does enhance parking success rate by 40.6%, improve parking completion efficiency by 18.0%, and expand ODD by 86.1%. It shows its superiority in difficult parking cases, such as the parallel parking scenario and narrow spaces. Moreover, the average computation time of the proposed planner is 74 milliseconds. Results indicate that the proposed planner is ready for real-time commercial applications.

## I. Introduction

In 2021, the global Automated Parking Assist (APA) system market is 10.1 billion USD [1]. The penetration rate of APA has reached a high level of 12.3% [2]. However, the user acceptance rate of APA is still quite low, as it is the third least-used application among all thirty-three Advanced Driver Assistance Systems (ADAS) functions, according to *Driver Interactive Vehicle Experience Report* [3]. There are two critical reasons. First, current APA systems are with low success rate, since they can hardly handle narrow spaces [4]. Second, current APA systems are time-consuming due to higher computation time and more "D-R" gear shiftings. Therefore, existing APA systems need to be further enhanced.

Generally, the APA system consists of three modules: perception, planning, and control. A centimeter-level perception has already been realized by high-precision sensors and environment recognition algorithms. Moreover, control methods are also with high accuracy [5]. State-of-the-art control error is less than 5cm [6, 7]. Therefore, perception and control are ready for the commercialization of APA [4]. The key to enhancing the existing APA system is the development of an APA planner.

There are mainly three types of parking planners, including geometry-based planners, search-based planners, and optimal-control-based planners [8, 9].

Geometry-based planners formulate paths by connecting the starting position with the target position through geometric lines [10]. Geometry-based planners consist of customized planners and fixed-library planners. Customized planners are proposed at the earliest, such as spline curve [11, 12]. However, they cannot consider vehicle dynamics. Sometimes it is impossible for the vehicle to trace the planned curve. It would decrease the execution success rate. To address this problem, fixed-library planners are proposed, such as reeds-sheep curve [13, 14]. The parking path is formulated by selecting and combining curves retrieved from a fixed library. All curves in the fixed library meet the requirement of vehicle dynamics. However, its planning domain is greatly limited because of its fixed nature. The shortcoming becomes critical especially in narrow spaces, as it can hardly plan a parking path. Therefore, the parking success rate of geometry-based planners is always a concern.

Search-based planners generate paths by searching a feasible path on a map [15, 16]. The map is divided into grids. Then the search process is conducted by exploring, picking, and connecting grid points. However, the computation efficiency reduces with the increase of environment complexity. In a complex environment, a smaller grid is needed to model the boundary of numerous irregular obstacles. It significantly increases the number of

*This paper is partially supported by National Science and Technology Major Project (No. 2022ZD0115501), National Natural Science Foundation of China (Grant No. 52302412 and 52372317), Yangtze River Delta Science and Technology Innovation Joint Force (No. 2023CSJGG0800), Shanghai Automotive Industry Science and Technology Development Foundation (No. 2404), Xiaomi Young Talents Program, the Fundamental Research Funds for the Central Universities, Tongji Zhongte Chair Professor Foundation (No. 000000375-2018082), the Science Fund of State Key Laboratory of Advanced Design and Manufacturing Technology for Vehicle (No. 32215011), the Postdoctoral Fellowship Program (Grade B) of China Postdoctoral Science Foundation (GZB20230519), Shanghai Sailing Program (No. 23YF1449600), Shanghai Post-doctoral Excellence Program (No.2022571), and China Postdoctoral Science Foundation (No.2022M722405). *(Corresponding author: Haoran Wang)*

Jia Hu, Yongwei Feng, and Shuoyuan Li, are with Key Laboratory of Road and Traffic Engineering of the Ministry of Education, Tongji University, Shanghai, China, 201804, (e-mail: hujia@tongji.edu.cn; 2210176@tongji.edu.cn; 2131304@tongji.edu.cn).

Haoran Wang is with Key Laboratory of Road and Traffic Engineering of the Ministry of Education, Tongji University, Shanghai, China, 201804, and the State Key Laboratory of Advanced Design and Manufacturing for Vehicle Body, Hunan University, Changsha, 410082, China (e-mail: wang_haoran@tongji.edu.cn).

Jaehyun So is with the Department of Transportation System Engineering, Ajou University, Suwon-si, Gyeonggi-do, 16499, Republic of Korea (e-mail: jso@ajou.ac.kr).

Junnian Zheng is with Hyperview Mobility (Shanghai) Co.,Ltd., No.488 Anchi Rd, Shanghai, 201805 (e-mail: junnian.zheng@hongjingdrive.com).

grids on the map and further increases search time [17]. Past studies show that even minutes are needed to find a parking path with search-based planners. Therefore, search-based planners can not ensure real-time application.

Optimal-control-based planners generate paths by solving optimization problems [18-22]. Zhang et al. [23] and Sheng et al. [24] generate parking paths by optimizing the coarse path from the search-based planners. Pagot et al. [25] generate parking paths by formulating a minimum-time optimal control problem with a novel collision avoidance design. They could output detailed motion commands, such as acceleration and steering angle, with the consideration of vehicle dynamics[26-28]. It enables the vehicle to precisely trace a planned curve. However, optimal-control-based planners are still concerned with the time-consuming problem due to frequent "D-R" gear shiftings [29]. To reach a target position in a narrow space, the final state error cost is given with a great weighting factor. It makes the planner short-sighted. The planned path would fast arrive at the vicinity of the target position and then oscillate around it [30], as shown in Fig. 1. Oscillating path leads to multiple driving direction switchings, which reduce driving comfortability and increase parking completion time. Besides, all current parking controllers lack a clear Operation Design Domain (ODD). A clear ODD is quite crucial for the commercialization of APA to understand where an APA system is functional [31].

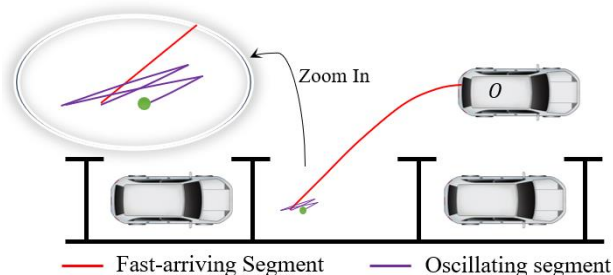

Fig. 1. Conventional parking path planning results

Therefore, a novel parking motion planner is proposed in this paper to address the above problems. It is highlighted for its logic: mirroring the parking target. Conventionally, a parking motion planner aims at generating paths towards the target parking slot. However, the driving direction switching may be a must, as illustrated in Fig. 2(a). The planning domain includes both reverse-and-forward driving. This significantly increases planning complexity, thereby difficultly finding optimal parking path. By proving the equivalence between reverse-driving and forward-driving, the reverse-and-forward planning can be transformed into a monodirectional planning by mirroring the parking target, as shown in Fig. 2(b). Therefore, the proposed motion planner can directly plan trajectory towards the mirrored parking target without considering driving direction switching. It significantly reduces planning complexity and enhances planning optimality.

Due to the highlighted mirroring parking target logic, this method has the following contributions:

• **Enhancing parking capability in narrow spaces**: Conventionally, APA system generally fast arrives at the vicinity of the target position and then adjust around the target position, as shown in Fig. 1. While, in a narrow slot, there may be not enough space for adjustment. Comparatively, by mirroring the parking target, the proposed motion planner directly optimizes trajectories toward the mirrored target with a globally planning vision. The first segment of reverse driving would make space for the forward driving adjustment, as shown in Fig. 2 (b). It would enhance the parking success rate in narrow spaces.

• **Ensuring a faster completion of parking**: The mirroring parking target logic is good at tackling planning with driving direction switching. The number of driving direction switching could be reduced. Frequent oscillations around the target position can be avoided. This ensures a faster parking completion.

• **Enhancing computational efficiency:** Based on the mirroring parking target logic, the planning domain could be reduced via transforming reverse-and-forward planning into monodirectional planning. A smaller optimization domain leads to faster solving. Furthermore, a solving algorithm is proposed to reduce the computation complexity caused by the integer variables in obstacle avoidance constraints. In our algorithm, the number of integer variables is significantly reduced by only adding constraints on identified steps within a prediction horizon.

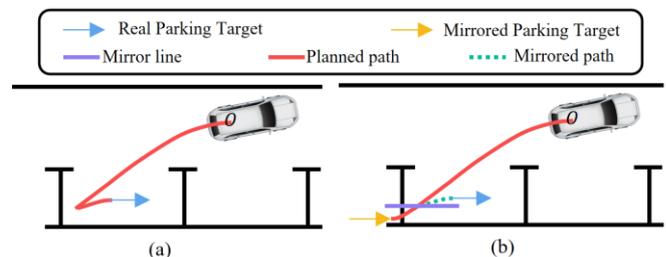

Fig. 2 Example of mirrored parking target

The remainder of this paper is organized as follows. Section II proposes the methodology of this paper. Section III verifies the proposed motion planner using simulation. Section IV draws a conclusion and provides future research discussion.

## II. METHODOLOGY

### A. Control Logic

The purpose of this paper is to propose an optimal-control-based motion planner. Scenarios of interest are illustrated in Fig. 3, including parallel, reverse, and angle parking scenarios.

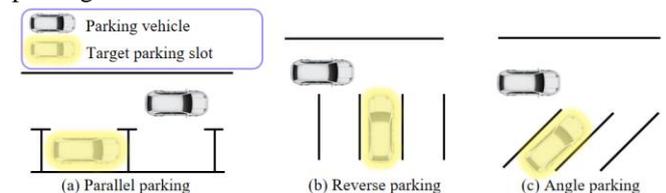

Fig. 3. Three types of parking

The parking system is presented in Fig. 4. It consists of 5 modules. Module 1 senses the surrounding environment. Module 2 determines the position of the mirror line. Module 3 makes planning decisions. Module 4 plans trajectories and outputs control commands. Module 5 is an execution layer. Details of this system are provided as follows:

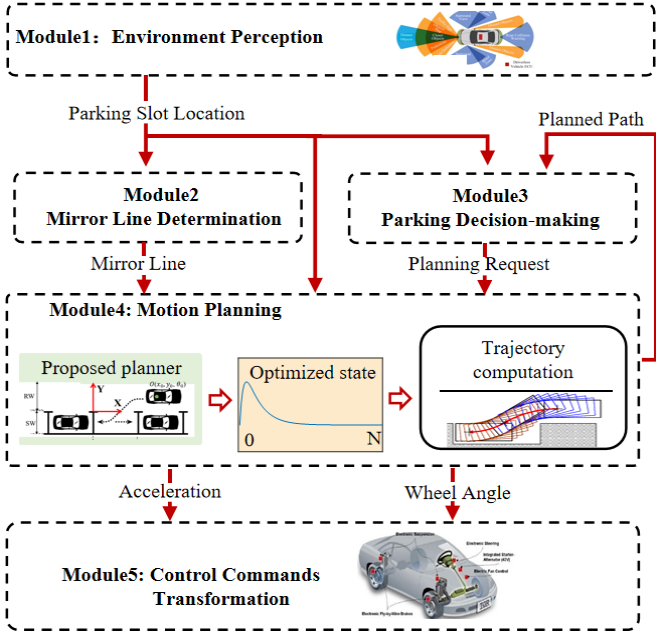

Fig. 4. The proposed APA system

• Module 1: This module collects the location of the parking slot.

• Module 2: The mirror line is determined in this module for all types of parking scenarios, as shown in Fig. 5. It is used for determining a mirrored parking target, so that reverse-and-forward planning could be transformed into monodirectional planning.

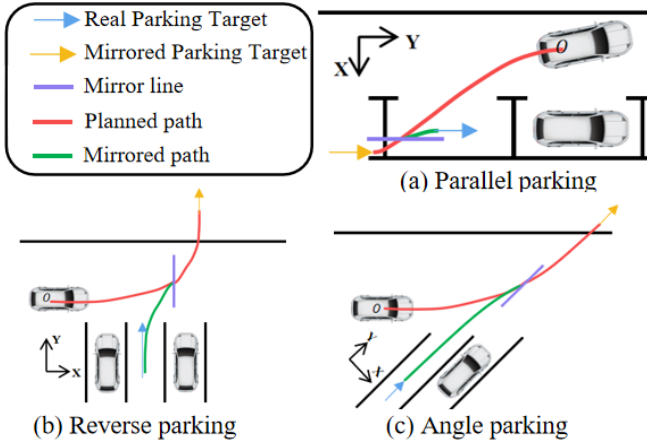

Fig. 5 Definition of the mirror line

• Module 3: This module conducts decision-making in the parking processes. It outputs planning requests which would activate Module 4.

• Module 4: This module plans parking trajectories towards the mirrored parking target. The planned motion commands, including acceleration and steering angle, are passed to module 5.

• Module 5: Control commands provided by Module 4 control the vehicle towards the mirrored parking target. This module transforms the control commands to control the vehicle towards the real parking target, as shown in Fig. 6. Motion commands before the mirror line is directly executed by local control. Motion commands after the mirror line is executed negatively by local control.

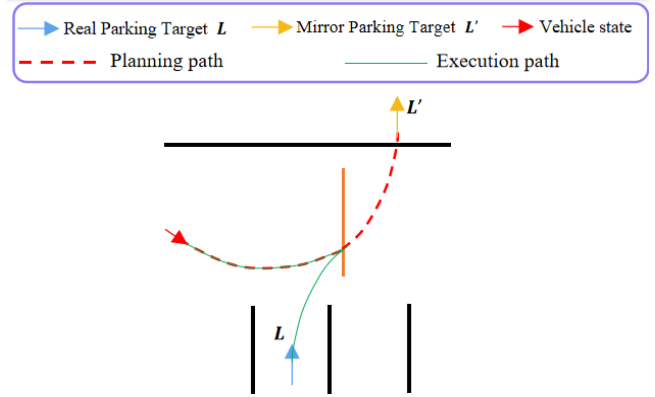

Fig. 6 Example of control commands transformation

TABLE I lists the indices and parameters utilized hereafter.

TABLE I Indices and parameters

| Parameter | Description |
|---|---|
| $a$ | Acceleration ($m/s^2$) |
| $a_{com}$ | Commanded acceleration ($m/s^2$) |
| $a_{min}$ | Minimum acceleration ($m/s^2$) |
| $a_{max}$ | Maximum acceleration ($m/s^2$) |
| $a_l^k$ | Coefficient of the $l_{th}$ edge of $R_{if}^k$. |
| $A$ | Coefficient matrix in vehicle dynamics model |
| $B$ | Coefficient matrix in vehicle dynamics model |
| $b_l^k$ | Coefficient of the $l_{th}$ edge of $R_{if}^k$. |
| $c_l^k$ | Coefficient of the $l_{th}$ edge of $R_{if}^k$. |
| $f$ | The full problem |
| $\varphi$ | Heading angle with respect to x direction ($rad$) |
| $i$ | Step in prediction horizon |
| $J$ | Cost function of the full problem |
| $K$ | The number of infeasible regions |
| $l$ | Distance between rear axle and front axle ($m$) |
| $l_1$ | Distance between rear axle and the rear most part of the vehicle ($m$) |
| $l_2$ | Distance between rear axle and the front most of the vehicle ($m$) |
| $l_3$ | Vehicle width ($m$) |
| $M$ | A large enough number in the big M method |
| $N$ | Total control steps |
| $Q$ | Weighting factor of state error in the motion planner |
| $r_{min}$ | Vehicle's minimum turning radius ($m$) |
| $R$ | Weighting factor of control penalty in the motion planner |
| $R_{if}^k$ | $k_{th}$ infeasible region |
| $RW$ | Road Width ($m$) |
| $SL$ | Length of parking slot ($m$) |
| $SW$ | Width of parking slot ($m$) |
| $t$ | Time (seconds) |
| $u$ | Control vector of vehicle |
| $v$ | Vehicle speed ($m/s$) |
| $v_{min}$ | Minimum vehicle speed ($m/s$) |
| $v_{max}$ | Maximum vehicle speed ($m/s$) |
| $x$ | Center of vehicle rear axle in x direction ($m$) |
| $y$ | Center of vehicle rear axle in y direction ($m$) |
| $x_f$ | Coordinate of real parking target in x direction ($m$) |
| $x_{f'}$ | Coordinate of mirrored parking target in x direction ($m$) |
| $x_{p^m}$ | Coordinate of $m_{th}$ feature point in x direction ($m$) |
| $y_{p^m}$ | Coordinate of $m_{th}$ feature point in y direction ($m$) |
| $z_l^k$ | Binary variable in big M method |
| $\beta$ | Tire slip angle ($rad$) |
| $\theta$ | Vehicle's heading angle ($rad$) |
| $\theta_f$ | Angle of real parking target ($rad$) |
| $\theta_{f'}$ | Angle of mirrored parking target ($rad$) |
| $\delta_f$ | Front-wheel angle ($rad$) |
| $\delta_{f_{com}}$ | Commanded front-wheel angle ($rad$) |
| $\delta_{f_{min}}$ | Minimum front-wheel angle ($rad$) |

| $\delta_{f\_max}$ | Maximum front-wheel angle ($rad$) |
|---|---|
| $\xi$ | State vector of vehicle |
| $\xi_{des}$ | Desired state vector of vehicle |
| $\Delta t$ | Discrete time step size (seconds) |
| $\tau_{\delta_f}$ | First-order inertia delay of steering (seconds) |
| $\tau_a$ | First-order inertia delay of acceleration (seconds) |
| $\Theta$ | A set of steps with collision avoidance constraints$\Theta$ |

*B. Mirror Line Determination*

In this section, a mirror line determination method is proposed for all types of parking scenarios.

*1) Parallel parking*

**Theorem 1**: Considering target reachability and collision avoidance, in parallel parking scenarios, the distance between the mirror line and the real parking target $l_{mi}$ shall be constrained by Eqs. (1) and (2).

$$l_{mi} \geq r_{min}(1 - \cos\theta) \tag{1}$$

$$l_{mi} \leq \frac{SW}{2} - l_1 \sin\theta - \frac{l_3}{2}\cos\theta \tag{2}$$

where $l_{mi}$ is the distance between the mirror line and the real parking target. $r_{min}$ is the vehicle's minimum turning radius. $\theta$ is the vehicle's heading angle. $l_1$ is the distance between rear axle and the rear end of the vehicle. $l_3$ is vehicle width.

**Proof:**

At the position of mirror line, the vehicle shall be able to reach the real parking target within its turning capacity, in order to ensure target reachability. Considering the vehicle's minimum turning radius is $r_{min}$ and the vehicle heading is $\theta$, Eq. (1) is obtained based on geometric relations, as shown in Fig. 7(a).

To ensure collision avoidance, the bottom corner of the vehicle (pink point) should not collide with the parking slot, when the vehicle arrives at the mirror line, as shown in Fig. 7(b). Based on geometric relations, $l$ is deduced as Eq. (3):

$$l = l_1 \sin\theta - \frac{l_3}{2}\cos\theta \tag{3}$$

To avoid collisions, Eq. (4) should be satisfied.

$$l + l_{mi} < \frac{SW}{2} \tag{4}$$

By substituting Eq. (3) into Eq. (4), Eq. (2) is obtained. This concludes the proof. ∎

Based on Theorem 1, the feasible domain of mirror position $l_{mi}$ is illustrated in Fig. 8. $l_{mi}$ is correlated with the vehicle heading angle. The blue line is the upper boundary. The red line is the lower boundary. The overlap is the feasible domain. Moreover, to ensure greater robustness on the heading angle, the dark point is suggested in practical applications.

*2) Reverse parking*

Simpler than parallel parking, the vehicle trajectory leading towards a reverse parking slot has only one requirement in the vicinity of the mirror line: when the vehicle is on the mirror line, it should be able to park into the slot directly within its turning capacity. The collision avoidance requirement no longer stands, as a vehicle would never bump into the boundary of a parking slot at the position of a mirror line. Hence, according to geometric relations, the distance between the mirror line and the real parking target $l_{mi}$ shall be constrained by Eq. (5), as shown in Fig. 9. A mirror line could be determined according to users' preferences as long as the constraint is satisfied.

$$l_{mi} \geq r_{min}(1 - \cos(\frac{\pi}{2} - \theta)) \tag{5}$$

where $l_{mi}$ is the distance between the mirror line and the real parking target. $r_{min}$ is the vehicle's minimum turning radius. $\theta$ is the vehicle's heading angle.

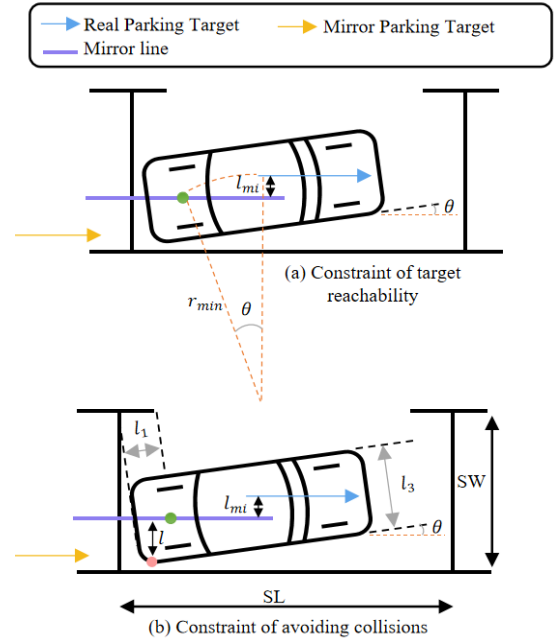

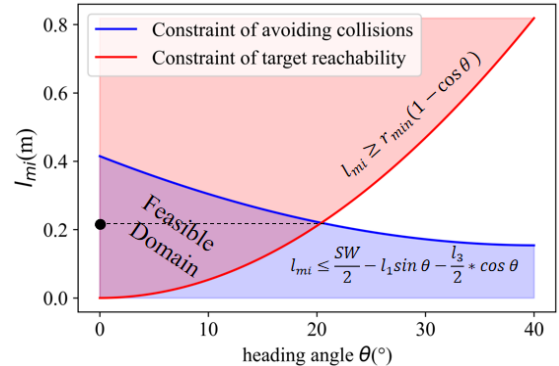

Fig. 7 Mirror line of parallel parking

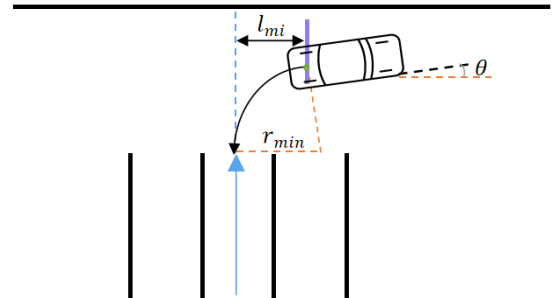

Fig. 8 Upper and lower boundaries of mirror line

Fig. 9 Mirror line of reverse parking

*3) Angle parking*

Similar to reverse parking, as shown in Fig. 10, the distance between the mirror line and the real parking target $l_{mi}$ shall be constrained by Eq. (6), to ensure target reachability.

$$l_{mi} \geq r_{min}(1 - cos(\frac{\pi}{4} - \theta)) \quad (6)$$

where $l_{mi}$ is the distance between the mirror line and the real parking target. $r_{min}$ is the vehicle's minimum turning radius. $\theta$ is the vehicle's heading angle.

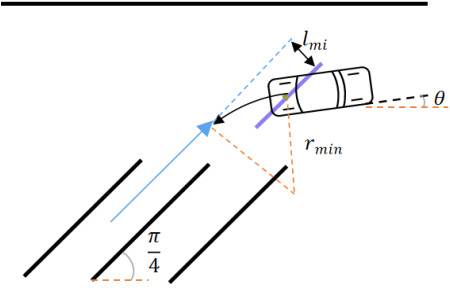

Fig. 10 Mirror line of angle parking

*C. Parking Decision-making*

In the parking maneuver, the vehicle may not park into the slot within one direction switching, especially in narrow spaces. To address the problem, a parking decision-making algorithm is proposed as shown in Fig. 11. At the initial state, a parking trajectory is planned with the objective of reaching the parking target. If the parking target cannot be realized, the planning is re-executed, starting from the end of last planning. After times of re-planning, vehicles could park into the slot.

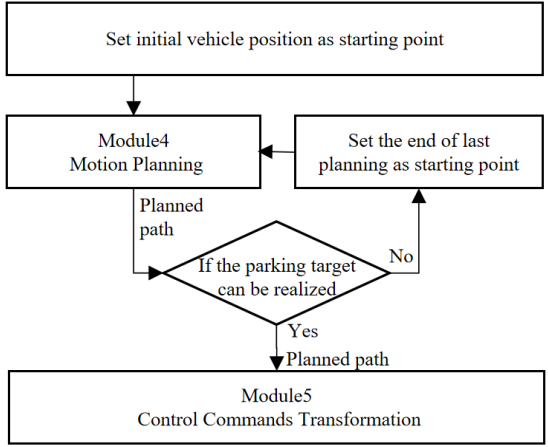

Fig. 11 Parking decision-making algorithm

*D. Motion Planning Problem Formulation*

The formulation of motion planning is presented in this section.

*1) Proof of equivalence between Reverse-driving and Forward-driving*

This section proves the equivalence between reverse-driving and forward-driving as shown in **Theorem 2**. By proving it, a reverse-and-forward planning can be transformed into a monodirectional planning.

**Theorem 2**: It is assumed that two vehicles are at the same position on mirror line and have opposite driving direction, as shown in Eq. (7) and Fig. 12. By executing opposite control commands in Eq. (8), the mirrored path of forward-driving vehicle is symmetric to the planned path of reverse-driving vehicle, as shown in Eq. (9).

$$\begin{cases} x_{back}(0) = x_{forth}(0) = 0 & (a) \\ y_{back}(0) = y_{forth}(0) = 0 & (b) \\ \theta_{back}(0) = \theta_{forth}(0) & (c) \\ v_{back}(0) = -v_{forth}(0) & (d) \end{cases} \quad (7)$$

where $x$, $y$, and $\theta$ represent the location and heading angle of the vehicle. $v$ represents the vehicle's speed. The subscript back and forth represent the reverse-driving vehicle and the forward-driving vehicle respectively. 0 represents the initial time.

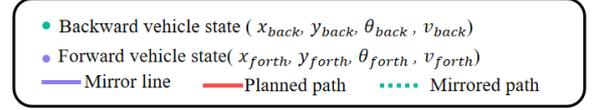
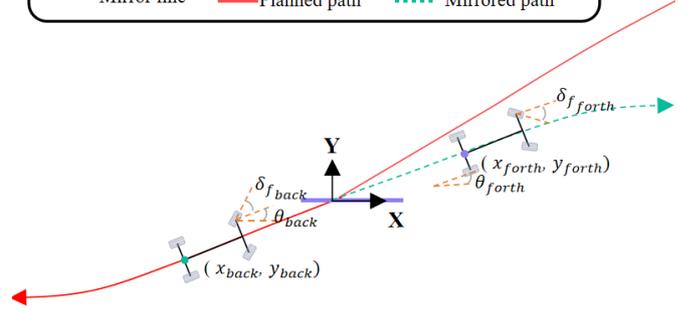

Fig. 12 Reverse-driving and Forward-driving vehicle

$$\begin{cases} \delta_{f\,forth}(t) = -\delta_{f\,back}(t) & (a) \\ a_{forth}(t) = -a_{back}(t) & (b) \end{cases} \quad (8)$$

$$\begin{cases} x_{back}(t) = -x_{forth}(t) & (a) \\ y_{back}(t) = -y_{forth}(t) & (b) \\ \theta_{back}(t) = \theta_{forth}(t) & (c) \\ v_{back}(t) = -v_{forth}(t) & (d) \end{cases} \quad (9)$$

**Proof:**

The vehicle dynamics are generally modeled by a kinematic bicycle model as follows [32].

$$\begin{cases} \dot{x} = v \cos \theta \\ \dot{y} = v \sin \theta \\ \dot{\theta} = \dfrac{v \tan \delta_f}{l} \\ \dot{v} = a \end{cases} \quad (10)$$

where $x$ and $y$ represent the center of the vehicle's rear axle. $v$ is the vehicle's speed. Its direction is the direction of the car body. $\theta$ is the heading angle with respect to $x$ direction. $l$ is the distance between rear axles and front axles. $\delta_f$ and $a$ are control commands. They are front-wheel angle and acceleration respectively.

Applying opposite control commands into Eq. (10), the states of the two vehicles propagate as shown in Eqs. (11) and (12):

$$\begin{cases} x_{forth}(t) = x_{forth}(0) + \int_0^t v_{forth}(\tau) \cos \theta_{forth}(\tau)\, d\tau & (a) \\ y_{forth}(t) = y_{forth}(0) + \int_0^t v_{forth}(\tau) \sin \theta_{forth}(\tau)\, d\tau & (b) \\ \theta_{forth}(t) = \theta_{forth}(0) + \int_0^t \dfrac{v_{forth}(\tau) \tan \delta_{f\,forth}(\tau)}{l}\, d\tau & (c) \\ v_{forth}(t) = v_{forth}(0) + \int_0^t a_{forth}(\tau)\, d\tau & (d) \end{cases} \quad (11)$$

$$\begin{cases} x_{back}(t) = x_{back}(0) + \int_0^t v_{back}(\tau) \cos\theta_{back}(\tau)\,d\tau & (a) \\ y_{back}(t) = y_{back}(0) + \int_0^t v_{back}(\tau) \sin\theta_{back}(\tau)\,d\tau & (b) \\ \theta_{back}(t) = \theta_{back}(0) + \int_0^t \dfrac{v_{back}(\tau)\tan\delta_{f_{back}}(\tau)}{l}\,d\tau & (c) \\ v_{back}(t) = v_{back}(0) + \int_0^t a_{back}(\tau)\,d\tau & (d) \end{cases} \quad (12)$$

Based on Eq. (8)(b), Eq. (13) is obtained.

$$\int_0^t a_{back}(\tau)\,d\tau = -\int_0^t a_{forth}(\tau)\,d\tau \quad (13)$$

By substituting Eq. (13) and Eq (7)(d) into Eq. (12)(d), Eq. (14) is obtained.

$$v_{back}(t) = -\left(v_{forth}(0) + \int_0^t a_{forth}(\tau)\,d\tau\right) \quad (14)$$

By substituting Eq. (11)(d) into Eq. (14), Eq. (15) is obtained.

$$v_{back}(t) = -v_{forth}(t) \quad (15)$$

Similarly, by substituting Eq (7), Eq. (8)(a), and Eq. (15) into Eq. (11) and Eq. (12), Eq. (16) is obtained.

$$\theta_{back}(t) = \theta_{forth}(t) \quad (16)$$

By substituting Eq. (15) and Eq. (16) into Eq. (11)(a-b) and Eq. (12)(a-b), Eq. (17) is obtained.

$$\begin{cases} \int_0^t v_{forth}(\tau)\cos\theta_{forth}(\tau)\,d\tau = -\int_0^t v_{back}(\tau)\cos\theta_{back}(\tau)\,d\tau \\ \int_0^t v_{forth}(\tau)\sin\theta_{forth}(\tau)\,d\tau = -\int_0^t v_{back}(\tau)\sin\theta_{back}(\tau)\,d\tau \end{cases} \quad (17)$$

By substituting Eq (7)(a-b) and Eq. (17) into Eq. (11)(a-b) and Eq. (12)(a-b), Eq. (18) is obtained.

$$\begin{cases} x_{back}(t) = -x_{forth}(t) & (a) \\ y_{back}(t) = -y_{forth}(t) & (b) \end{cases} \quad (18)$$

By combining Eq. (15), Eq. (16), and Eq. (18), Eq. (9) is obtained.

It shows that the mirrored path of forward-driving vehicle is symmetric to the planned path of reverse-driving vehicle. Therefore, Theorem 2 is concluded. ∎

*2) State Definition*

State and control vectors are defined in **Definition 1** and **Definition 2.**

**Definition 1**: Vehicle state vector $\boldsymbol{\xi}$ and control vector $\boldsymbol{u}$ are defined as follows:

$$\boldsymbol{\xi} \overset{\text{def}}{=} [x(i), v(i), a(i), y(i), \theta(i), \delta_f(i)]^T \quad (19)$$

$$\boldsymbol{u} \overset{\text{def}}{=} [a_{com}(i), \delta_{f_{com}}(i)]^T \quad (20)$$

**Definition 2**: Desired vehicle state vector $\boldsymbol{\xi}_{des}$ are defined as follows.

$$\boldsymbol{\xi}_{des} \overset{\text{def}}{=} [x_{f'}, 0, 0, \sim, \theta_{f'}, 0]^T \quad (21)$$

where $x_{f'}$ is the coordinate of the mirrored parking target in x direction. $\theta_{f'}$ is the heading angle of the mirrored parking target.

*3) System Dynamics*

Bicycle model is used as the vehicle dynamics [33-35]. Small yaw angle assumption is adopted when linearize vehicle dynamics. To avoid large yaw angle, local linearization is conducted around the heading of parking trajectory. A linearizing angle is designed close to the average yaw angle of parking path, as shown in Fig. 13.

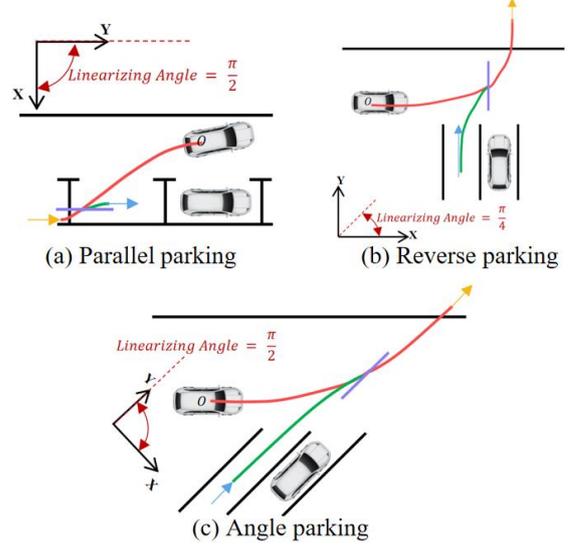

(a) Parallel parking (b) Reverse parking

(c) Angle parking

Fig. 13 Linearizing angle of local linearization

The linearized vehicle dynamics are as follows.

$$x_{i+1} = x_i + v_i \gamma_1 \Delta t \quad (22)$$

$$y_{i+1} = y_i + v_i \gamma_2 \Delta t \quad (23)$$

$$\theta_{i+1} = \theta_i + v_i \frac{\delta_{f_i}}{l} \Delta t \quad (24)$$

$$v_{i+1} = v_i + a_i \Delta t \quad (25)$$

where $l$ is the distance between rear axles and front axles. $\Delta t$ is the discrete time step size. $\gamma_1$ and $\gamma_2$ are coefficients related to linearizing angle. They are defined as follows.

$$\gamma_1 = -\theta_i \sin Angle + \cos Angle + Angle \sin Angle \quad (26)$$

$$\gamma_2 = \theta_i \cos Angle + \sin Angle - Angle \cos Angle \quad (27)$$

where $Angle$ is the linearizing angle for three types of parking.

To model the local control response accurately, vehicle control delay is considered by a first-order inertia model as follows:

$$a_{i+1} = a_i + \frac{a_{com_i} - a_i}{\tau_a} \Delta t \quad (28)$$

$$\delta_{f_{i+1}} = \delta_{f_i} + \frac{\delta_{f_{com_i}} - \delta_{f_i}}{\tau_{\delta_f}} \Delta t \quad (29)$$

where $\tau_{\delta_f}$ and $\tau_a$ are the first-order inertia delay of steering and acceleration respectively.

By applying the Eqs. (22)-(29) into Eqs. (19)-(20), vehicle dynamics models are formulated as follows.

$$\boldsymbol{\xi}_{i+1} = \boldsymbol{A}_i \boldsymbol{\xi}_i + \boldsymbol{B}_i u_i \quad (30)$$

$$A_i = \Delta t \times \begin{bmatrix} 0 & \gamma_1 & 0 & 0 & 0 & 0 \\ 0 & 0 & 1 & 0 & 0 & 0 \\ 0 & 0 & \frac{-1}{\tau_a} & 0 & 0 & 0 \\ 0 & \gamma_2 & 0 & 0 & 0 & 0 \\ 0 & 0 & 0 & 0 & 0 & \frac{v}{l} \\ 0 & 0 & 0 & 0 & 0 & \frac{-1}{\tau_{\delta_f}} \end{bmatrix} + I_{6\times 6} \quad (31)$$

$$B_i = \Delta t \times \begin{bmatrix} 0 & 0 \\ 0 & 0 \\ 1 & 1 \\ \tau_a & \tau_{\delta_f} \end{bmatrix} \quad (32)$$

*4) Cost Function*

To accelerate computation, cost functions are both formulated into a quadratic form. The objective function is formulated as follows:

$$\min_u \left( \sum_{i=1}^{N} \left( \underbrace{Q\|\xi_i - \xi_{des}\|_2}_{\text{State error cost}} + \underbrace{R\|u_i\|_2}_{\text{Control cost}} \right) + \underbrace{Q\|\xi_{N+1} - \xi_{des}\|_2}_{\text{Terminal state error cost}} \right) \quad (33)$$

where $Q\|\xi_i - \xi_{des}\|_2$ is the state error cost. $R\|u_i\|_2$ is control effort cost. Weighting factors of state error cost $Q$ is a positive-definite matrix. Weighting factors of control effort cost $R$ is also a positive-definite matrix.

*5) Constraints*

*a) State and Control Constraint*: Vehicle speed should be constrained within speed limits. Vehicle's acceleration should be bounded by vehicle capability and comfort. Vehicle's front wheel angle should be within its steering range [36, 37].

$$v_{min} \le v \le v_{max} \quad (34)$$

$$a_{min} \le a \le a_{max} \quad (35)$$

$$\delta_{f\,min} \le \delta_f \le \delta_{f\,max} \quad (36)$$

where $v_{min}$ and $v_{max}$ are the minimum and maximum vehicle speed respectively. $a_{min}$ and $a_{max}$ are the minimum and maximum acceleration respectively. $\delta_{f\,min}$ and $\delta_{f\,max}$ are minimum and maximum front-wheel angle respectively.

*b) Constraints on Vehicle State on the Mirror Line:* The vehicle would switch its driving direction at the position of the mirror line. Therefore, the speed must be zero on the mirror line.

$$v_i = 0 \quad if\,|x_i - l_1| \le \varepsilon \quad (37)$$

where $\varepsilon$ is a small positive number.

*c) Obstacles Avoidance Constraint*: To avoid collisions, the ego vehicle should be outside all infeasible parking regions $R_{if}^k$, as shown in Fig. 15. Six feature points are selected to represent the ego vehicle, including, as shown in Fig. 14, where $l_1$ is the distance between rear axle and the rear most part of the vehicle. $l_2$ is the distance between rear axle and the front most of the vehicle. $l_3$ is vehicle width. For six feature points, including $P_1(x_{p_i^1}, y_{p_i^1})$, $P_2(x_{p_i^2}, y_{p_i^2})$, $P_3(x_{p_i^3}, y_{p_i^3})$, $P_4(x_{p_i^4}, y_{p_i^4})$, $P_5(x_{p_i^5}, y_{p_i^5})$, and $P_6(x_{p_i^6}, y_{p_i^6})$, their coordinates can be easily computed by geometric relations. For example, $x_{p_i^1}$ equals $x_i + l_2 \cos\theta_i - \frac{l_3}{2}\sin\theta_i$.

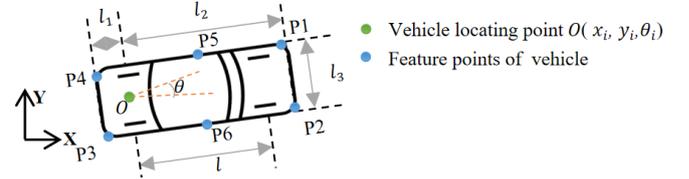

Fig. 14. Illustration of notations

The obstacle avoidance constraint could be formulated as follows.

$$\left( x_{p_i^m}, y_{p_i^m} \right) \notin R_{if}^k \quad (38)$$
$$\forall m \in \{1,2,3,4,5,6\}\ \forall k \in \{1,2,\ldots,K\}$$

where $p_i^m$ is $m_{th}$ feature point. $R_{if}^k$ is the $k_{th}$ convex infeasible region as shown in Fig. 15.

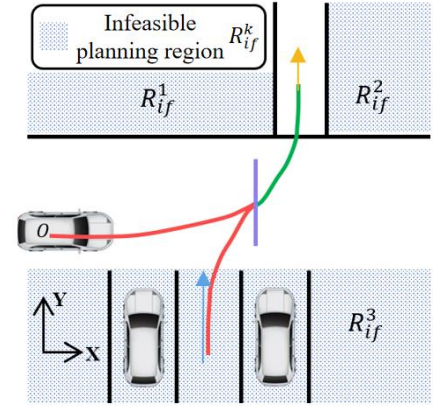

Fig. 15. Obstacles avoidance example

For the $k_{th}$ convex infeasible region $R_{if}^k$, it can be represented by linear inequalities as shown in Eq. (39)

$$R_{if}^k \overset{\text{def}}{=} \bigcap_{l=1}^{L} a_l^k x + b_l^k y + c_l^k \le 0 \quad (39)$$

where $l$ represents the $l_{th}$ edge of $R_{if}^k$. $a_l^k$, $b_l^k$ and $c_l^k$ are coefficients of the $l_{th}$ edge of $R_{if}^k$.

By adopting the big M method [38], Eq. (38) is formulated as follows.

$$\bigcap_{l=1}^{L} a_l^k x_{p_i^m} + b_l^k y_{p_i^m} + c_l^k \ge M(1 - z_l^k)$$
$$\sum_{l=1}^{L} z_l^k \ge 1 \quad (40)$$
$$\forall m \in \{1,2,\ldots,6\}\ \forall k \in \{1,2,\ldots,K\}\ \forall i \in \Theta$$

where $M$ is a large enough number. $z_l^k$ is a binary variable. $\Theta$ is a set that records which control steps need obstacle avoidance constraints.

*6) Full problem*

The full problem $f$ is formulated as follows.
$$J \to \text{Eq.}(33)$$
s.t.
$$\text{system dynamics} \to \text{Eqs.}(30) - (32) \quad (41)$$
state and control constraint
$$\to \text{Eqs.}(34) - (36)$$

mirror line constraint → Eq. (37)
obstacles avoidance constraint → Eq. (40)

where $J$ is the cost function of $f$.

By mirroring parking target, the full problem can directly plan trajectory from current state towards the mirrored parking target. The optimality within once driving direction switching is ensured.

*7) Problem Solving*

The proposed optimal-control-based motion planning problem is transformed into a Mixed Integer Quadratic Programming (MIQP) problem and can be solved by Gurobi optimizer [39]. Since binary variables would decrease computational efficiency, an iterative solving algorithm is adopted in this research.

As presented by ALGORITHM I, an initial trajectory is generated by solving a QP problem without considering collision avoidance. After that, the key steps where collision happens are identified by finding the intersection between the initial trajectory and the infeasible region $R_{if}$. These key steps are the steps where collision avoidance constraints shall be added in the next iteration. The MIQP problem in the next iteration only consists of a small number of binary variables and could be solved efficiently. After times of iteration, the feasible and sub-optimal trajectory could be found. Furthermore, the problem solving does not need a lot times of iteration, since there are only simple and static obstacles in our problem.

ALGORITHM I Solving algorithm

**Input**: $R_{if} = \bigcup_{k=1}^{K} R_{if}^{k}, \forall k \in \{1,2,...,K\}$
**Initialize**: the set of steps with collision avoidance constraints $\Theta = \emptyset$
**Output**: optimal parking trajectory $P^*$
get $f$ according to Eq. (41) except Eq. (40)
$P^* = QP(f)$
**While** $P^* \cap R_{if} \neq \emptyset$ **then**
  **For** $i \in \{0,1,...,N\}$ **then**
    **If** $P^*(i) \cap R_{if} \neq \emptyset$ **then**
      $\Theta = \Theta \cup i$
    **End if**
  **End For**
  use $\Theta$ to add collision avoidance constraints to $f$ according to Eq. (40)
  $P^* = MIQP(f)$
**End while**

*E. Control Commands Transformation*

Based on Theorem 2, to control the vehicle towards the real parking target, the planned motion commands after the mirror line is executed oppositely, as shown in Eq. (42).

$$\begin{cases} \delta_{f_{executed}}(t) = -\delta_{f_{planned}}(t) & (a) \\ a_{executed}(t) = -a_{planned}(t) & (b) \end{cases} \quad (42)$$

where $\delta_{f_{planned}}$ and $a_{planned}$ is the planned control command. $\delta_{f_{executed}}$ and $a_{executed}$ is the executed control command.

## III. EVALUATION

The proposed parking motion planner is evaluated from the following five aspects: i) function validation; ii) parking capability quantification; iii) parking reliability evaluation; iv) parking completion efficiency evaluation; v) computation efficiency evaluation.

*A. Experiment Design*

The experiment is conducted on a simulation platform previously developed by this research team [18, 40]. The simulation platform is also coupled with Carsim, to simulate the high-fidelity vehicle dynamics. The following settings are adopted, as shown in TABLE II.

TABLE II Parameter settings

| Parameter | Description | Settings |
|---|---|---|
| $a_{min}$ | Minimum acceleration ($m/s^2$) | -5 |
| $a_{max}$ | Maximum acceleration ($m/s^2$) | 3 |
| $l$ | Distance between rear axle and front axle ($m$) | 2.5 |
| $l_1$ | Distance between rear axle and the rear most part of the vehicle ($m$) | 0.71 |
| $l_2$ | Distance between rear axle and the front most part of the vehicle ($m$) | 3.11 |
| $l_3$ | Vehicle width ($m$) | 1.67 |
| $N$ | Total control steps | 60 |
| $v_{min}$ | Minimum vehicle speed ($m/s$) | -3 |
| $v_{max}$ | Maximum vehicle speed ($m/s$) | 3 |
| $\delta_{f_{min}}$ | Minimum front-wheel angle ($rad$) | -0.6 |
| $\delta_{f_{max}}$ | Maximum front-wheel angle ($rad$) | 0.6 |
| $\Delta t$ | Control step-size (seconds) | 0.1 |
| $\tau_{\delta_f}$ | First-order inertia delay of steering (seconds) | 0.1 |
| $\tau_a$ | First-order inertia delay of acceleration (seconds) | 0.3 |

*1) Test Scenarios*

The proposed motion planner is tested in three types of parking scenarios including parallel parking, reverse parking, and angle parking, as shown in Fig. 16. The vehicle is in a space comprised of slots and road. The slot size is determined by two factors: Slot Length (SL) and Slot Width (SW). Road Width (RW) indicates the distance between the parking slot and the road boundary. The vehicle would like to park into the slot from its initial position. The vehicle's initial position is defined by three factors: initial heading $\theta_0$, initial distance to slot $y_0$, and initial distance to road boundary $y_1$.

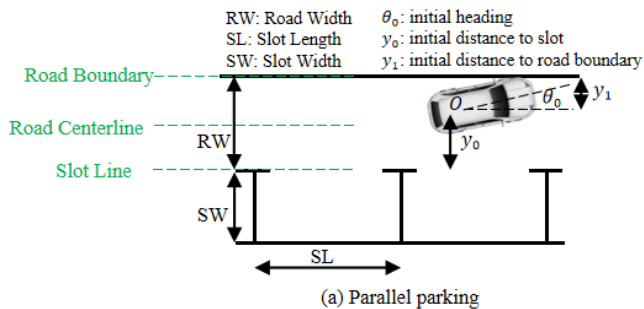
(a) Parallel parking
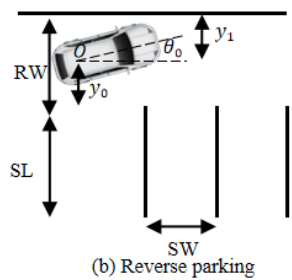
(b) Reverse parking
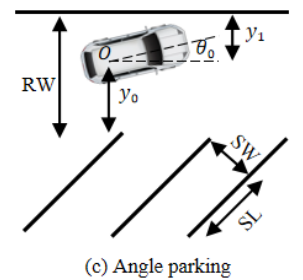
(c) Angle parking

Fig. 16. Sensitive factors

*2) Motion Planner Types*

Two types of motion planners are evaluated:

• **Baseline planner**: This planner is a search-based planner [35, 41]. It is short-sighted due to its rationale that the planning is always towards the parking target.

• **The proposed planner**: This planner is optimal control based. It enhances parking performance due to its global optimality.

*3) Measures of Effectiveness*

The following Measures of Effectiveness (MOEs) are adopted.

*a) Function Validation*

Planned trajectories are used to verify the proposed function. The function of interest is mirroring the parking target.

*b) Parking Capability in Narrow Spaces Quantification*

The success rate of parking is used to quantify parking capability in narrow spaces. Criteria of a successful parking maneuver are defined according to ISO 16787 [42] and ISO 20900 [43] as follows.

*i) Criteria of Successful Parallel Parking:* As illustrated in Fig. 17, successful parallel parking requires that:

• Heading angle error θ is between -3° and 3°;

• $M_f$, $M_r$, and $M_e$ are all greater than 0;

• Parking time duration must be less than 180 seconds;

• No collisions;

• The vehicle outline is inside the parking slot.

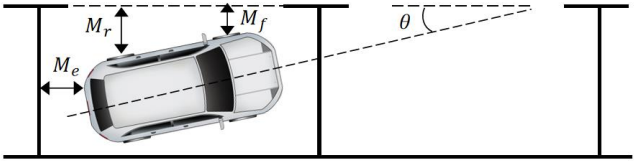

Fig. 17. Parallel parking example

*ii) Criteria of Successful Reverse and Angle Parking:* As illustrated in Fig. 18, successful reverse and angle parking require that.

• Heading angle error θ is between -3° and 3°;

• $M_{fl}$, $M_{fr}$, $M_{rl}$, $M_{rr}$, and $M_e$ are all greater than 0.1m;

• Parking time duration must be less than 180 seconds;

• No collisions;

• The vehicle outline is inside the parking slot.

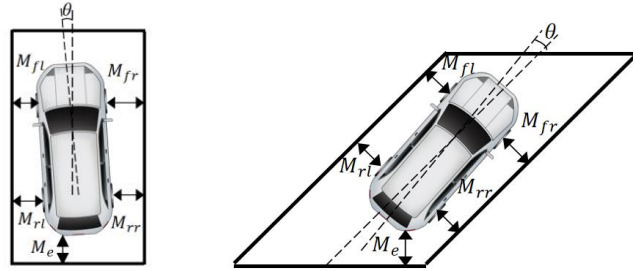

(a) Reverse parking    (b) Angle parking
Fig. 18. Reverse and angle parking example

*c) Parking reliability Evaluation*

ODD is used to evaluate parking reliability. ODD is a collection of initial conditions which could successfully complete the parking process with a rate of 95%. Initial conditions consist of the following factors: Slot Length (SL), Slot Width (SW), initial heading $\theta_0$, and initial distance to slot $y_0$. A larger ODD indicates a more reliable parking planner.

*d) Parking Completion Efficiency Evaluation*

Parking completion efficiency is quantified by two measurements: the parking time duration and the number of driving direction switching [44].

*e) Computation Efficiency Evaluation*

The computation time of parking path planning is used to evaluate computation efficiency.

*4) Sensitivity Analysis*

The sensitivity analysis is conducted in terms of space size and initial position. Six factors in Fig. 16 are selected, including Road Width (RW), Slot Length (SL), Slot Width (SW), vehicle's initial heading $\theta_0$, vehicle's initial distance to slot $y_0$, and vehicle's initial distance to road boundary $y_1$. Levels of factors are set in TABLE III, according to ISO 16787 (*Assisted parking system — Performance requirements and test procedures*) [42] and ISO 20900 (*Partially automated parking systems — Performance requirements and test procedures*) [43]. By iterating through factor levels and excluding the cases unable to be generated, there are 71,009 cases, including 12995 parallel parking cases, 46629 reverse parking cases, and 11385 angle parking cases.

TABLE III Cases Design

| Types of Parking | Road Width (RW) | Parking slot size | | Initial heading $\theta_0$ | Initial distance to slot $y_0$ |
| --- | --- | --- | --- | --- | --- |
| | | Length (SL) | Width (SW) | | |
| Parallel parking | 4.5m | Minimum: 3.82m | 2.5m | Minimum: -90deg Maximum: 90deg Sampling step: 10deg | Minimum: 0m Maximum: Road Width Sampling step: 0.1m |
| | 4.0m | Maximum: 7.35m | | | |
| | 3.5m | Sampling step: 0.1m | | | |
| Reverse parking | 7m | 4.82m | Minimum: 1.67m Maximum: 3.27m Sampling step: 0.05m | | |
| | 6m | | | | |
| | 5m | | | | |
| Angle parking | 4.5m | | | | |
| | 4m | | | | |
| | 3.5m | | | | |

## B. Results and Discussions

Results demonstrate that the proposed motion planner is: i) with the function of mirroring the parking target; ii) enhancing parking success rate by 40.6%; iii) expanding ODD by 86.1%; iv) reducing 47.6% number of driving direction switching and 18.0% parking time duration; v) with 74 milliseconds of average computation time.

## 1) Function Validation Results

The function of mirroring the parking target is verified in Fig. 19. It illustrates the planned trajectories and the real parking trajectories in all three types of parking scenarios. The planned trajectories are towards a mirrored parking target. By executing negative control commands after the mirror line, the vehicle can park into the slot successfully.

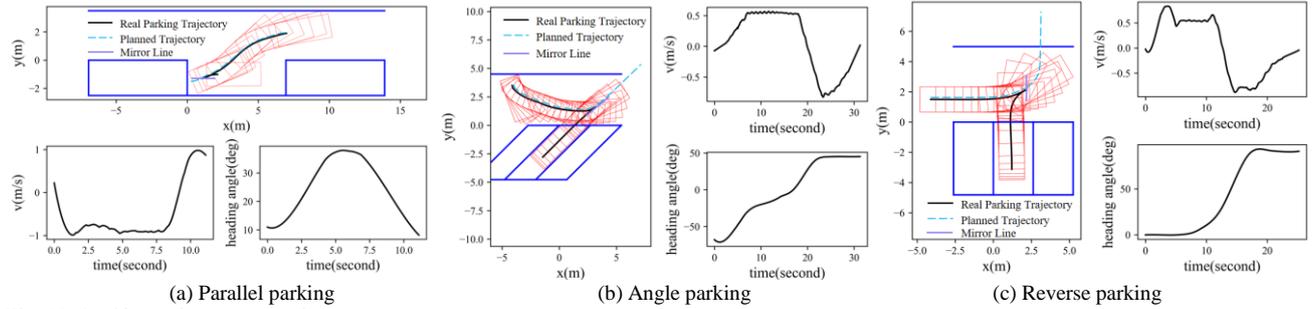

(a) Parallel parking  (b) Angle parking  (c) Reverse parking

Fig. 19. Parking trajectory examples

## 2) Parking Capability in Narrow Spaces Results

The proposed planner is able to enhance parking capability by 40.6%, compared to the baseline controller. The parking capability is quantified by parking success rate, as illustrated in Fig. 20. It shows that the proposed motion planner is with greater parking success rate in all parking types.

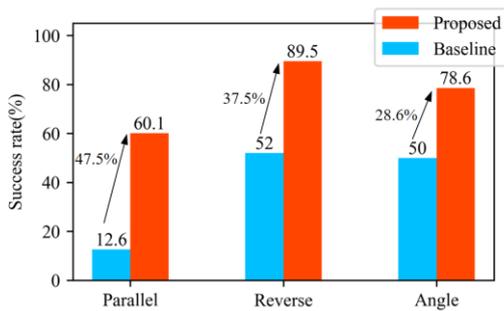

Fig. 20. Success rate results

The proposed planner particularly shows its superiority in more difficult cases. First, the proposed planner has greater enhancement on parking capability in difficult parking types, like parallel parking [45]. As shown in Fig. 20, the parking success rate enhancement in parallel parking is 47.5%. It is significantly greater than the enhancement in reverse parking and angle parking. Second, the proposed planner has greater enhancement on parking capability in narrower spaces. As shown in Fig. 21, the parking success rate enhancement increases with the decrease of road width. It is because the proposed planner barely deteriorates with the decrease of road width, as shown in Fig. 21. It is able to take extreme measures to find a way to squeeze into the parking slot. An example trajectory is provided in Fig. 22. In this case, the proposed planner made over many direction switches, and finally accomplished its parking process. While, the baseline planner attempts to search for a path directly leading towards the parking slot. Due to its greedy nature, all the attempts failed as the shorter paths are more likely constrained by the narrower spaces.

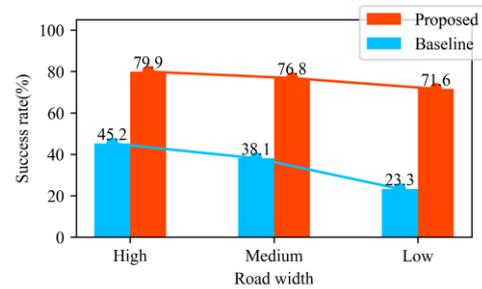

Fig. 21 Success rate VS road width

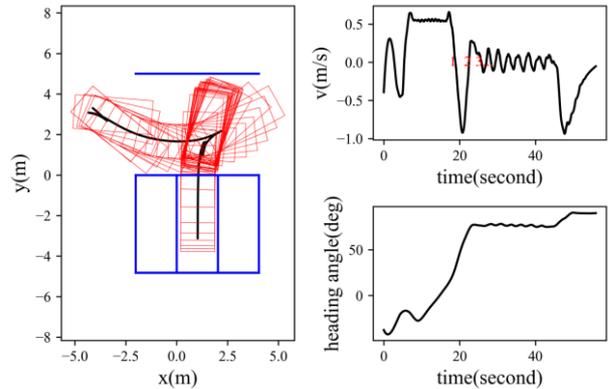

Fig. 22 Example of parking capability in narrow spaces

The result of sensitivity analysis of parking capability in terms of slot size is shown in the column I of Fig. 23. It demonstrates that the proposed planner enhances parking capability. The enhancement is particularly significant with a narrower parking slot. The magnitude of the enhancement becomes increasingly significant as slot size decreases. This confirms the proposed planner's parking capability in narrow spaces. It is discovered that the baseline planner cannot ensure a 100% parking success rate regardless of space sizes in parallel parking. This is in line with daily experiences that parallel parking is the most difficult parking type. For people, that are able to handle reverse parking and angle parking, cannot always complete a parallel parking. Nevertheless, the proposed planner can handle parallel parking as good as the other two parking types. This confirms the robustness of the proposed planner.

It is interesting to find that the success rate of the conventional planner does not increase linearly with slot size. Take angle parking as an example, the success rate remains constant when the slot width is around 2 meters. It is due to the nature of the baseline planner, as it needs to complete the final path by connecting towards the parking slot with a predetermined curve. Since the curve is selected from a fixed library, the degree of freedom is limited around the final path. Hence, there are chances that two adjacent positions could have two significantly different success rates. This causes the success rate not always continuously increase with slot size. This is not the case for the proposed planner. By mirroring the parking target, the proposed planner is able to perform global optimizations and make the most of the space available. Therefore, any increment in slot size leads to a gain in success rate. This again confirms the superiority of the proposed planner when dealing with narrow parking slots.

The result of sensitivity analysis of parking capability in terms of vehicle's initial position is shown in the column II and III of Fig. 23. Result shows that parking success rate increases with the free space around the initial position. It is confirmed by the column II of Fig. 23 that success rate is higher when the vehicle's initial position is farther from the slot. It is confirmed again by the column III of Fig. 23 that success rate is higher when the vehicle's initial position is farther from the road boundary. However, the success rate does not always increase with the space available. It reaches the maximum success rate of around 100% when the distance to slot is over 1.6 meters, as there is simply no room for further improvement. Therefore, to achieve the best user experience, future users are suggested to activate the system after placing their vehicles at least 1.6 meters away from the slot.

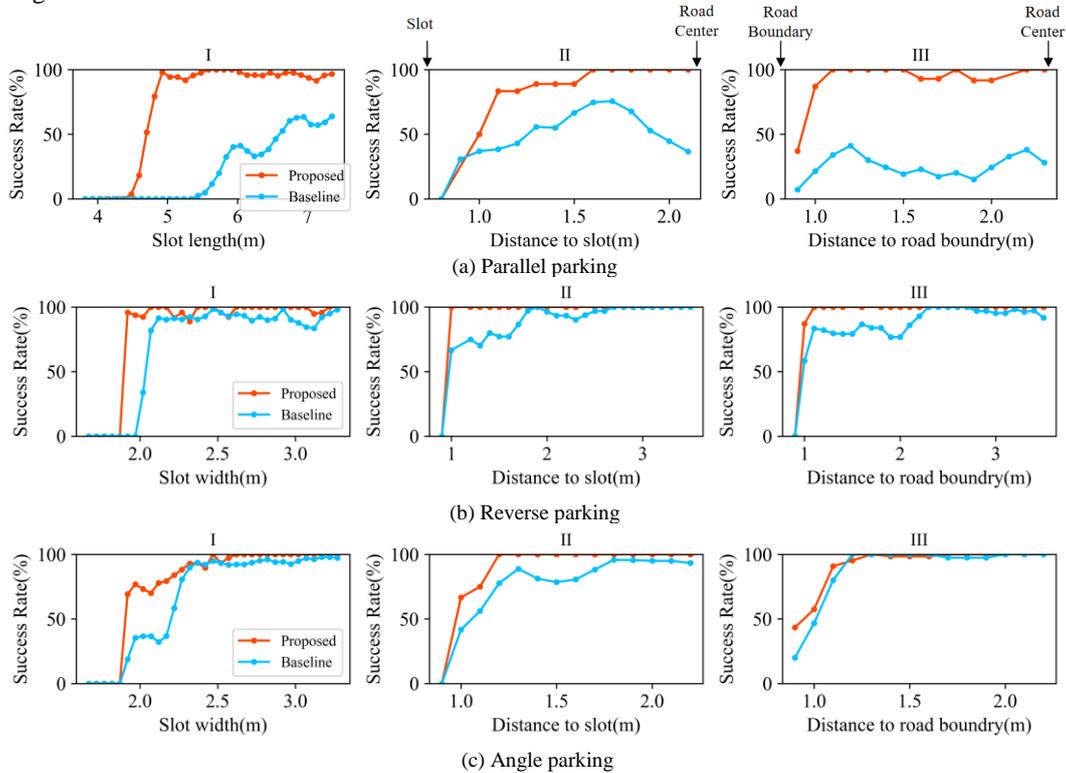

Fig. 23. Sensitivity analysis for parking capability

*3) Parking reliability Results*

The proposed planner is confirmed with greater parking reliability. Parking reliability is quantified by ODD, as shown in Fig. 24. Green nodes are the successful cases. Red nodes are the failed cases. ODD is the collection of green nodes. It shows that the proposed planner expands the ODD by 86.1%, compared to the baseline planner. Therefore, the proposed planner is able to handle a greater variety of parking slots and conditions. This indicates its greater commercial implementation potential.

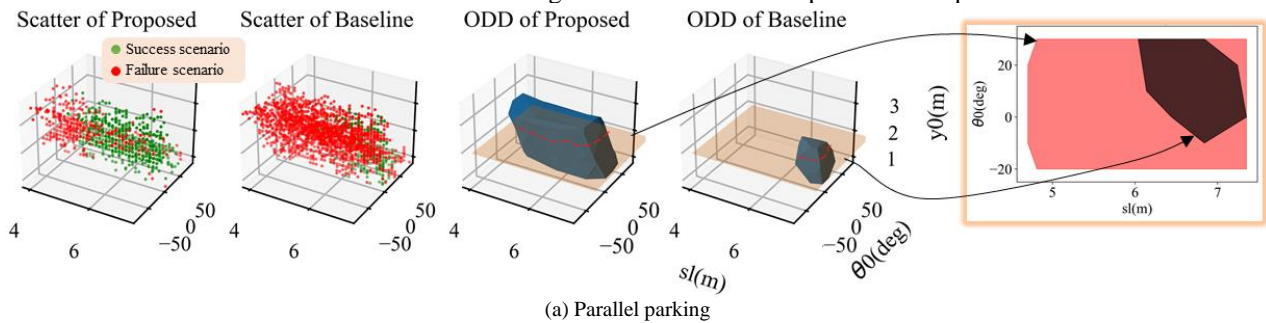

(a) Parallel parking

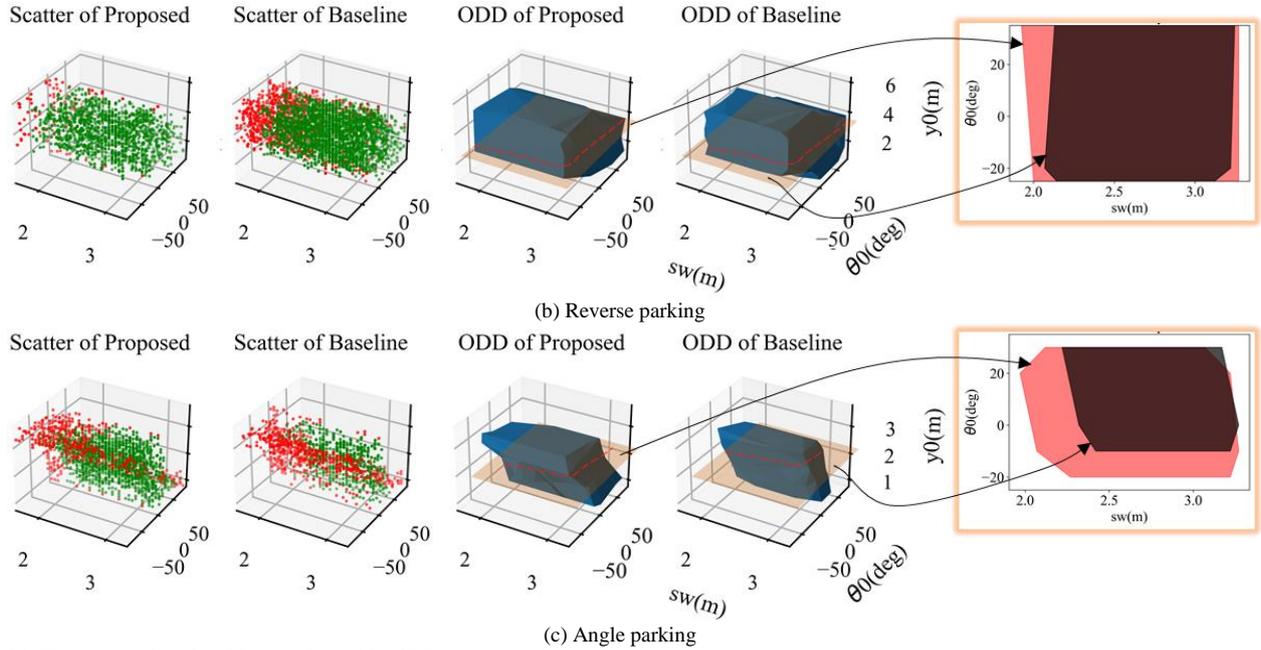

(b) Reverse parking

(c) Angle parking

Fig. 24. The scatter plot of parking results and its ODD representation

*4) Parking Completion Efficiency Results*

The proposed motion planner is with greater parking completion efficiency compared to the baseline. The parking completion efficiency has been quantified by the number of driving direction switching and parking time duration, as illustrated in Fig. 25. As shown by the left plots in Fig. 25, the proposed planner reduces 47.6% direction switchings. The average direction switching number is only 3.3. As shown by the right plots of Fig. 25, the proposed planner reduces parking time duration by 18.0%. It needs 28.2 seconds on average to finish a parking.

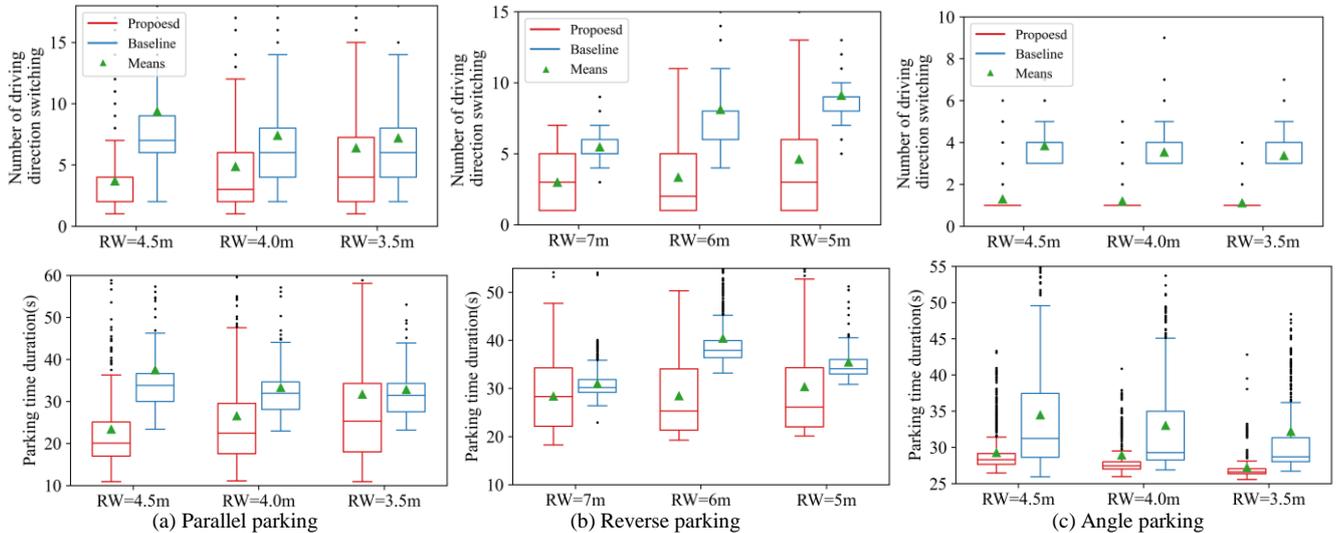

(a) Parallel parking  (b) Reverse parking  (c) Angle parking

Fig. 25. Parking performance results

*5) Computation Efficiency Results*

The proposed planner enhances computation efficiency by 95.8% compared to the baseline. When running on a laptop equipped with an Intel i7-9750H CPU, the average computation time of the proposed motion planner is only 74 milliseconds. While, the average computation time of the baseline planner is 1.8s. Consequently, the proposed motion planner is with the potential of real-time implementation.

## IV. CONCLUSION AND FUTURE RESEARCH

This research proposes an optimal-control-based parking motion planner. Its highlight lies in its control logic: planning trajectories by mirroring the parking target. This method enables: i) parking capability in narrow spaces; ii) better parking reliability by expanding ODD; iii) faster completion of parking process; iv) enhanced computational efficiency; v) universal to all types of parking. A comprehensive evaluation is conducted by simulation. Results demonstrate that:

•Compared to the conventional method, the proposed planner enhances parking optimality by enhancing parking success rate by 40.6%, improving parking completion efficiency by 18.0%, and expanding ODD by 86.1%;

•The proposed planner shows its superiority in difficult parking cases, such as the parallel parking scenario and narrow spaces;

- Users are advised to activate the proposed system after placing their vehicles at least 1.6 meters away from the slot. In this way, an 100% success rate could be guaranteed;

- The average computation time of the proposed planner is 74 milliseconds. It indicates that the proposed planner is ready for real-time application.

In this research, parking completion efficiency is the priority. Future research could consider and balance more parking objectives.

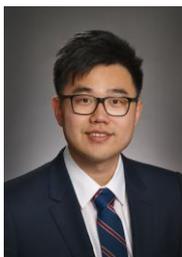
**Jia Hu** (Member, IEEE) is currently working as the Zhongte Distinguished Chair of Cooperative Automation with the College of Transportation Engineering, Tongji University. Before joining Tongji University, he was a Research Associate with the Federal Highway Administration (FHWA), USA. Furthermore, he is also a member of TRB (a division of the National Academies) Vehicle Highway Automation Committee, Freeway Operation Committee, and Simulation subcommittee of Traffic Signal Systems Committee, and a member of CAV Impact Committee and Artificial Intelligence Committee of the ASCE Transportation and Development Institute. He is also an Associate Editor of the American Society of Civil Engineers Journal of Transportation Engineering and an Assistant Editor of the Journal of Intelligent Transportation Systems, and an Advisory Editorial Board Member of the Transportation Research Part C: Emerging Technologies. He has been an Associate Editor of the IEEE Intelligent Vehicles Symposium since 2018 and the IEEE Intelligent Transportation Systems Conference since 2019.

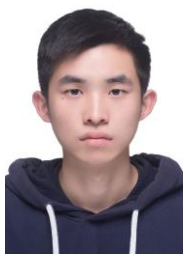
**Yongwei Feng** was born in Henan, China. He received the B.S. degree in traffic engineering from Tongji University in 2020, where he is currently pursuing the PHD's degree. His research interests include connected and automated vehicle, optimal control theory, and ITS.

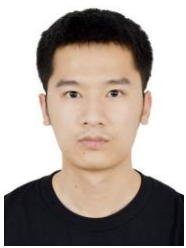
**Shuoyuan Li** was born in Shanxi, China. He received the B.S. degree in traffic engineering from Jilin University in 2020, where he is currently pursuing the master's degree. His research interests include transit signal and optimal control theory.

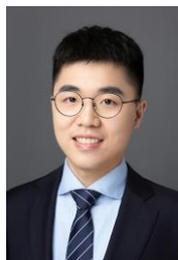
**Haoran Wang** received the bachelor's degree in transportation engineering from Tongji University, Shanghai, China, in 2017, and the Ph.D. degree from Tongji University in 2022. He is currently a postdoctoral researcher with the College of Transportation Engineering, Tongji University. He is a researcher on vehicle engineering, majoring in intelligent vehicle control and cooperative automation. Dr. Wang served the IEEE TRANSACTIONS ON INTELLIGENT VEHICLES, *Journal of Intelligent Transportation Systems*, and IET *Intelligent Transport Systems* as peer reviewers with a good reputation.

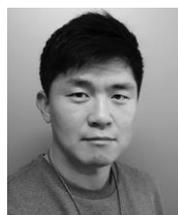
**Jaehyun (Jason) So** received the B.S. degree and the M.S. degree in transportation engineering from Ajou University, South Korea, in 2006 and 2008, respectively, and the Ph.D. degree in civil and environmental engineering from the University of Virginia, VA, USA, in 2013. He is currently an Assistant Professor with Department of Transportation System Engineering at Ajou University. His research interests include traffic operations, safety, and connected-automated vehicles.

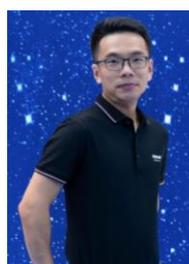
**Junnian Zheng** was born in Shanghai, China. He holds B.S. degree in mechanical engineering from Shanghai Jiao Tong University, and M.S. and Ph.D. degrees in mechanical engineering from Texas A&M University. He is currently the director of innovation and advanced engineering at Hyperview Mobility (Shanghai). His research interests include ADAS and autonomous driving system for passenger and commercial vehicles, embodied AI, and large language models for self-driving.